\documentclass[conference]{IEEEtran}
\IEEEoverridecommandlockouts
\usepackage{placeins}
\usepackage{cite}
\usepackage{amsmath,amssymb,amsfonts}
\usepackage{algorithmic}
\usepackage{textcomp}
\usepackage{xcolor}
\newcommand{\rev}[1]{{#1}} 
\usepackage{graphicx}
\usepackage{booktabs}

\newcommand{\zPD}{normProDTW}
\newcommand{\B}[1]{\textbf{#1}}   

\begin{document}

\title{Constructing Multiscale Pattern Dictionaries for Financial Time Series\\with a Regularized Procrustes--DTW Divergence}

\author{
\IEEEauthorblockN{Koutarou Tamura}
\IEEEauthorblockA{\textit{Nomura Research Institute, Ltd.}\\
\textit{Financial Technology Solutions Division}\\
Tokyo, Japan\\
k.tamura.phd@gmail.com, k9-tamura@nri.co.jp}
\and
\IEEEauthorblockN{Jun-ichi Maskawa}
\IEEEauthorblockA{\textit{Seijo University}\\
Tokyo, Japan\\
maskawa@seijo.ac.jp}
}

\maketitle

\begin{abstract}
Financial time series exhibit recurring local shapes that differ in duration and amplitude.
\rev{We propose normProDTW, a divergence that combines amplitude normalization, dynamic time warping (DTW), vertical flipping, and regularized Procrustes superimposition, and use it to construct, without supervision, a pattern dictionary of observed curves shared across timeframes from 1 to 1,440 one-minute observations.}
\rev{On market series built from FTSE 100 and Nikkei 225 constituents, the dictionary reconstructs an unseen test period more accurately on average than dictionaries based on DTW, Euclidean distance, or a k-Shape-based distance, with greater inter-element dissimilarity, and matches the accuracy of amplitude-normalized DTW with 50 elements using approximately 17--28 elements.}
\rev{Each occurrence records its element, orientation, timeframe, duration, and range, giving an inspectable discrete description of price paths that could be used by language models to generate market commentary.}
\end{abstract}


\section{Introduction}\label{sec:intro}

Price series contain local shapes---a reversal after a rise, a double bottom, oscillating consolidation---that recur across time scales.
Technical analysis names some of these shapes and uses them for trading decisions, and Lo et al.~\cite{lo2000foundations} detected hand-defined patterns algorithmically and examined them statistically.
Other work feeds price-chart images into convolutional networks to extract predictive signals without naming patterns~\cite{jiang2023reimagining}.

However, methods that define pattern conditions in advance miss unnamed shapes, and their conditions are often designed with a particular time scale in mind.
Similarity search and motif discovery in time-series data mining~\cite{faloutsos1994fast,rakthanmanon2012searching,yeh2016matrixprofile} and shape-based clustering~\cite{paparrizos2015kshape} find shapes in data, but they aim at search, discovery, or classification at a single length, not at describing a whole series with a few shapes across many timeframes.
Representations learned by deep models on images do not necessarily provide explicit, interpretable curve shapes.
Hence a compact, inspectable \emph{dictionary} of shapes that is extracted from data and shared across timeframes remains underexplored.
\rev{Such a dictionary is also relevant to applying language models to financial time series, for which a discrete representation of continuous prices can be useful. Chronos~\cite{ansari2024chronos} quantizes individual values; a shape dictionary instead uses a curve over a period as the unit, which can be described in words.}

Occurrences of the same pattern differ in duration, range, trend slope during formation, and noise, so building such a dictionary requires a measure of whether two curves have the same shape while accounting for these differences.
DTW~\cite{sakoe1978dtw} absorbs misalignment along the time axis but treats the same shape with different ranges as different patterns.
Amplitude normalization removes differences in level and range, but the same shape formed under different trends still differs in slope.
Procrustes analysis~\cite{gower1975procrustes} superimposes point sets by scaling, rotation, and translation, but it assumes known point correspondences, and without limits on its degrees of freedom it may judge any pair of curves similar.

We propose a divergence that combines point correspondence by DTW with penalized Procrustes superimposition (Section~\ref{sec:method}), and use it to construct, without supervision, a dictionary of cumulative-return curves that appear across timeframes and window lengths in 1-minute market series of FTSE~100 and Nikkei~225 constituents (Section~\ref{sec:data}).
The dictionary is evaluated by how accurately it reconstructs series in an unseen test period with few elements, and by the dissimilarity among its elements (Section~\ref{sec:design}).
\rev{We show that the proposed dictionary achieves higher average reconstruction accuracy and inter-element dissimilarity than the compared methods, and matches the reconstruction accuracy of normalized DTW with fewer elements (Section~\ref{sec:results}).}
\rev{This paper constructs and evaluates the representation itself; it does not evaluate language models or forecasting.}

\section{Related Work}\label{sec:related}

\subsection{Shape Distances and Superimposition}
DTW~\cite{sakoe1978dtw} and its extensions~\cite{keogh2001derivative,zhao2018shapedtw} accommodate warping of the time axis, methods for uniform temporal scaling~\cite{fu2008scaling} accommodate that deformation, and in the value direction, subsequence normalization~\cite{rakthanmanon2012searching} and complexity correction~\cite{batista2014cid} are used.
All of them, however, treat time and value separately and do not absorb differences in slope caused by trends.
Procrustes analysis~\cite{gower1975procrustes,dryden2016statistical} superimposes point sets by similarity transforms but assumes known correspondences.
Canonical time warping~\cite{zhou2009canonical} alternates DTW and linear projections to align multivariate series, but it seeks a common feature space between series and does not penalize the size of the transform.
Our method alternates DTW with a similarity transform on the time--value plane and includes the size of the transform in the divergence.

\subsection{Discovering and Representing Recurring Shapes}
Similarity search~\cite{faloutsos1994fast,rakthanmanon2012searching}, symbolic representation~\cite{lin2003sax}, motif discovery~\cite{yeh2016matrixprofile,linardi2018valmod}, and extraction of discriminative subsequences (shapelets)~\cite{ye2009shapelets} aim, respectively, at efficient search, symbolization of series, discovery of recurring subsequences, and extraction of shapes useful for classification;
none primarily aims to reconstruct series at different timeframes with a small set of shared shapes.
k-Shape~\cite{paparrizos2015kshape} computes centroids by shape-based clustering, DBA~\cite{petitjean2011dba} computes average series from DTW alignments, and soft-DTW~\cite{cuturi2017softdtw} enables optimization of average series through a differentiable alignment cost; in all of them the representative is a synthesized series and not necessarily an observed shape. \rev{We compare with the distances of Euclidean clustering and k-Shape within our own dictionary construction (Section~\ref{sec:methods}).}
\rev{Autoencoders~\cite{hinton2006reducing} learn latent representations whose variables do not necessarily correspond to observed curves.}
We use observed curves themselves as representatives (k-medoids~\cite{kaufman1990finding}) and build a dictionary shared across timeframes.
As in piecewise linear approximation~\cite{keogh2001online}, we evaluate the dictionary by the trade-off between approximation error and the number of elements.

\section{Data}\label{sec:data}

We use two market series constructed from 1-minute stock prices on the London Stock Exchange and the Tokyo Stock Exchange, with the same constituents, periods, and preprocessing as Maskawa~\cite{maskawa2025fluctuation}.
Let $P_{j,t}$ be the price of stock $j$ at time $t$ and $\tilde r_{j,t}=\log P_{j,t}-\log P_{j,t-1}$.
The return of each stock is standardized by its mean $\mu_j$ and standard deviation $\sigma_j$ and averaged over $N$ stocks:
\begin{equation}
r_t=\frac{1}{N}\sum_{j=1}^{N}\frac{\tilde r_{j,t}-\mu_j}{\sigma_j}.
\label{eq:market-return}
\end{equation}
This aligns the scales of individual stocks to extract the common market component; it is not a tradable index.
\rev{Because only stocks listed throughout each period are used, the series may contain survivorship bias.}
Intraday seasonality is removed before analysis.

\begin{figure}[tb]
\centering
\includegraphics[width=\linewidth]{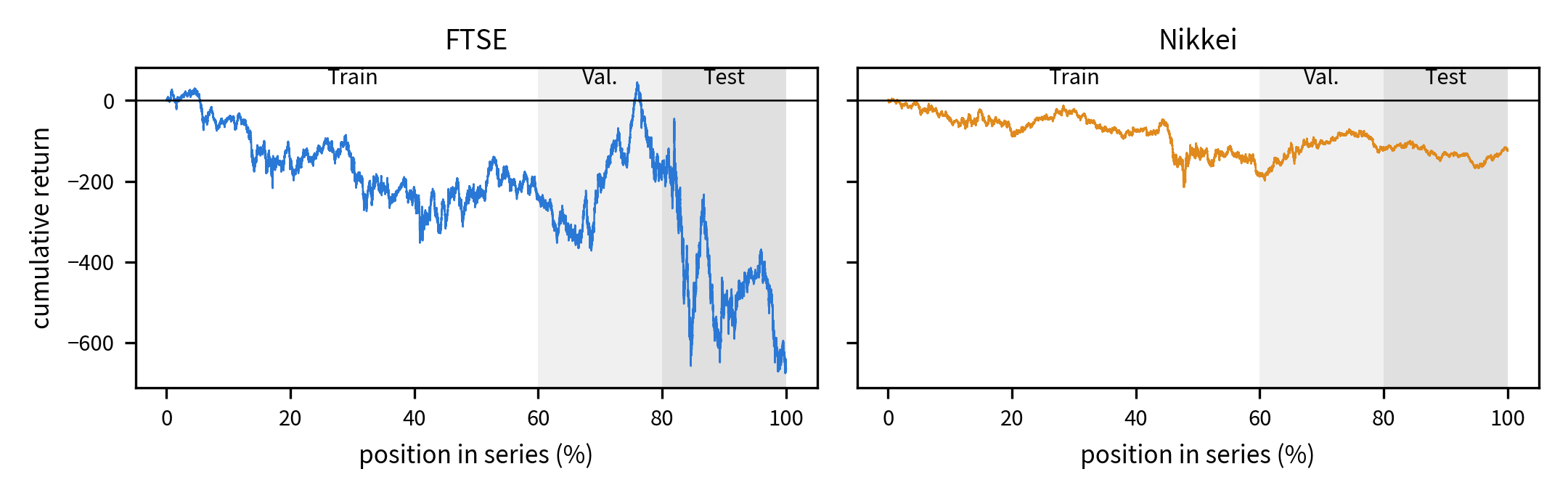}
\caption{Cumulative market-return series of FTSE and Nikkei with the Train/Validation/Test boundaries.}
\label{fig:data}
\end{figure}

The FTSE series is built from FTSE~100 constituents listed continuously from November 2007 to January 2009, and the Nikkei series from Nikkei~225 constituents from October 2007 to December 2009.
Both periods include the collapse of Lehman Brothers and the crash of October 2008.
Because overnight and lunch-break changes are removed, timeframes are counted in observations, and \texttt{1D} denotes 1,440 observations.
Thus \texttt{1D} is not a daily close-to-close return but a timeframe aggregating 1,440 one-minute observations.
Fig.~\ref{fig:data} shows the cumulative series, and Table~\ref{tab:data} gives summary statistics of the returns; both series are heavy-tailed.
Each series is split chronologically into 60\% Train, 20\% Validation, and 20\% Test, and windows straddling split boundaries are discarded.

\begin{table}[tb]
\centering
\caption{Summary statistics of 1-minute market returns}
\label{tab:data}
\small
\begin{tabular}{lrr}
\toprule
Statistic & FTSE & Nikkei \\
\midrule
Observations & 198,490 & 131,283 \\
Mean & $-0.0033$ & $-0.0010$ \\
Std.\ dev. & 1.004 & 0.390 \\
Skewness & $-0.098$ & $-0.217$ \\
Excess kurtosis & 12.60 & 12.80 \\
\bottomrule
\end{tabular}
\end{table}

\section{Proposed Method}\label{sec:method}

\begin{figure*}[!t]
\centering
\includegraphics[width=\textwidth]{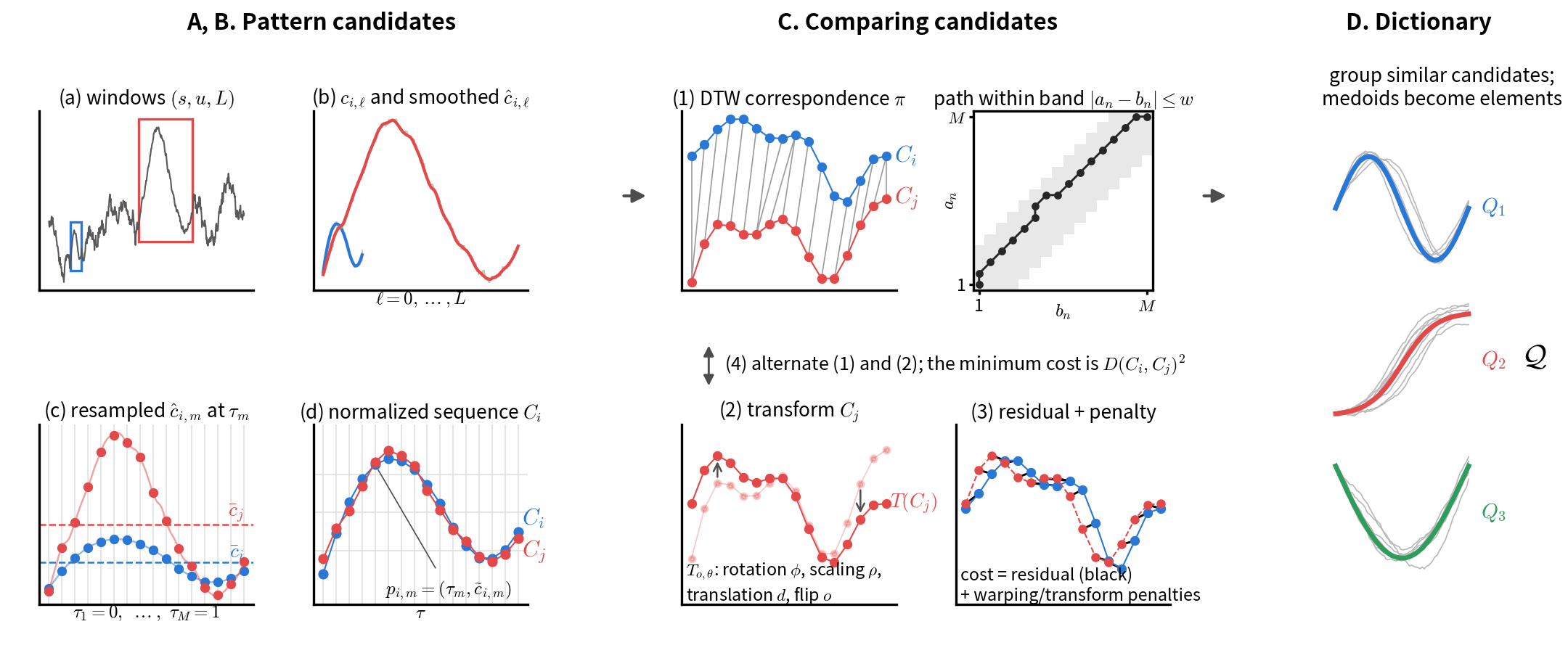}
\caption{Overview of dictionary construction (drawn with a smaller $M$). A, B: windows of different timeframes and lengths are extracted (a), smoothed (b), resampled at $M$ points (c), and normalized to form a candidate pattern $C_i$ (d) (Sections~\ref{sec:candidates} and~\ref{sec:znorm}). C: the banded DTW correspondence $\pi$ (1) and the regularized similarity transform (2) are updated alternately; \rev{the mean squared residual plus transform penalties is the cost (3), and the smaller final cost over the two orientations defines $D^2$} (Section~\ref{sec:distance}). D: similar candidate patterns are grouped, and the observed curve $Q_k$ representing each group becomes a dictionary element (Section~\ref{sec:dictionary}).}
\label{fig:framework}
\end{figure*}

The construction consists of extracting and normalizing candidate patterns (Sections~\ref{sec:candidates} and~\ref{sec:znorm}), comparing curves (Section~\ref{sec:distance}), and selecting representative observed curves as the dictionary (Section~\ref{sec:dictionary}) (Fig.~\ref{fig:framework}).

\subsection{Multiscale Candidate Patterns}\label{sec:candidates}
For timeframe $s$, the 1-minute returns~\eqref{eq:market-return} are summed over $f_s\in\{1,5,15,30,60,240,1440\}$ consecutive observations to give the return $r^{(s)}_t$ at time $t$.
A candidate pattern $i=(s,u,L)$ is defined by the timeframe $s$, start time $u$, and window length $L\in\mathcal L_s$ (Table~\ref{tab:settings}), and its cumulative-return curve $c_{i,\ell}$ is
\begin{equation}
c_{i,0}=0,\qquad c_{i,\ell}=\sum_{j=0}^{\ell-1}r^{(s)}_{u+j}\quad(\ell=1,\dots,L),
\label{eq:cumulative}
\end{equation}
a cumulative-return path starting from zero.

To compare curves of different lengths, $c_{i,\ell}$ is smoothed by the Savitzky--Golay filter~\cite{savitzky1964smoothing} to obtain $\hat c_{i,\ell}$, the formation time $\ell/L$ is mapped to $[0,1]$, and the values $\hat c_{i,m}$ at $M$ equally spaced points $\tau_m=(m-1)/(M-1)$ ($m=1,\dots,M$) are obtained by linear interpolation.
Candidate patterns of all timeframes and window lengths can thus be compared as cumulative-return curves of the same $M$ points (Fig.~\ref{fig:framework}(a)--(c)).

\subsection{Amplitude Normalization of Curves}\label{sec:znorm}
\rev{Similar shapes can differ in amplitude and vertical offset, for example between high- and low-volatility regimes.}
We therefore take the mean and standard deviation of the resampled curve $\hat c_{i,m}$,
\begin{equation}
\bar c_i=\frac{1}{M}\sum_{m=1}^{M}\hat c_{i,m},\qquad
s_i=\Bigl(\frac{1}{M}\sum_{m=1}^{M}(\hat c_{i,m}-\bar c_i)^2\Bigr)^{1/2},
\end{equation}
normalize the amplitude (level and range) of the curve as
\begin{equation}
\tilde c_{i,m}=\frac{\hat c_{i,m}-\bar c_i}{s_i},\qquad p_{i,m}=(\tau_m,\tilde c_{i,m})^\top
\label{eq:znorm}
\end{equation}
(Fig.~\ref{fig:framework}(d)), and represent candidate pattern $i$ by the planar point sequence $C_i=(p_{i,1},\dots,p_{i,M})$.
Hereafter this amplitude normalization is simply called normalization, and methods that apply it are prefixed with ``norm.''

\subsection{Regularized Procrustes--DTW Divergence}\label{sec:distance}
Between normalized curves $C_i$ and $C_j$, misalignment in formation timing and differences in slope due to trends still remain.
We therefore determine point correspondences by DTW, superimpose one curve on the other by scaling, rotation, and translation so that corresponding points come close, evaluate the residual alignment error together with penalties on the size of the transform, and update the correspondence and the transform alternately (Fig.~\ref{fig:framework}C).
Without constraints, the transform could make dissimilar curves appear similar, so the size of the transform is penalized and included in the divergence.

\subsubsection{Point correspondence}
Let the sequence of corresponding pairs be $\pi=((a_1,b_1),\dots,(a_N,b_N))$.
It is a monotone sequence from $(1,1)$ to $(M,M)$, and the time offset is restricted to $|a_n-b_n|\le w$ (Sakoe--Chiba band).

\subsubsection{Similarity transform}
The transform applied to a point $p=(\tau,\tilde c)^\top$ of $C_j$ is
\begin{equation}
T_{o,\theta}(p)=\rho\,R(\phi)\begin{pmatrix}\tau\\ o\,\tilde c\end{pmatrix}+d,
\end{equation}
where $\theta=(\rho,\phi,d)$ consists of the scale, rotation angle, and translation, $R(\phi)$ is the rotation matrix, and $o\in\{-1,+1\}$ includes vertical flipping so that mirror images of rises and falls are regarded as the same shape.
For fixed $\pi$, the unpenalized optimal scale $\rho_0$ and rotation $\phi_0$ are obtained in closed form by Procrustes analysis.
To avoid excessive transforms, the transform is defined by the following shrinkage rule, which moves them toward the identity with weights $\lambda$:
\begin{equation}
\rho=\rho_0^{1/(1+\lambda_s)},\qquad \phi=\frac{\phi_0}{1+\lambda_r},\qquad d=\frac{d_0}{1+\lambda_t},
\label{eq:shrink}
\end{equation}
where $d_0$ is the translation that matches the centroids of corresponding points under the shrunk $\rho$ and $\phi$.
Equation~\eqref{eq:shrink} is not the exact minimizer of the cost below but an approximation that moves the Procrustes solution toward the identity.

\subsubsection{Cost}
With the transform fixed, the correspondence $\pi$ is chosen by banded DTW, which minimizes the cumulative cost along the path by dynamic programming,
\begin{equation}
\sum_{n=1}^{N}\Bigl[\bigl\lVert p_{i,a_n}-T_{o,\theta}(p_{j,b_n})\bigr\rVert_2^2
+\lambda_\pi\Bigl(\frac{a_n-b_n}{M-1}\Bigr)^2\Bigr],
\label{eq:dtwcost}
\end{equation}
where the path length $N$ depends on $\pi$.
After the transform is determined, the residual is normalized by the path length, and the cost is
\begin{align}
E(\pi,o,\theta)=\;&\frac{1}{N}\sum_{n=1}^{N}\bigl\lVert p_{i,a_n}-T_{o,\theta}(p_{j,b_n})\bigr\rVert_2^2 \nonumber\\
&+\lambda_s(\log\rho)^2+\lambda_r\phi^2+\lambda_t\lVert d\rVert_2^2.
\label{eq:energy}
\end{align}
The time-offset penalty $\lambda_\pi$ is used only for choosing the correspondence and is not included in $E$.
The weights $\lambda$ are listed in Table~\ref{tab:settings}; they were fixed in advance and not tuned on Test results.

\subsubsection{Alternating updates}
Choosing the correspondence by~\eqref{eq:dtwcost} and updating the transform by~\eqref{eq:shrink} are alternated until the change in $E$ becomes sufficiently small or 10 iterations are reached.
Because~\eqref{eq:shrink} is an approximation, $E$ is not guaranteed to decrease at every update, and the value at termination is used.
\rev{This is done for each of $o=\pm1$, and the squared divergence between the two candidate patterns is the smaller final cost:}
\begin{equation}
D(C_i,C_j)^2=\min_{o\in\{-1,+1\}}E(\pi_o,o,\theta_o),
\label{eq:zpd}
\end{equation}
Here $\pi_o$ and $\theta_o$ are the values at the end of the alternating updates for orientation $o$; since the updates may stop at a local solution, \rev{$D^2$} is not necessarily the global minimum of $E$.
Because the transform is applied to $C_j$, $D$ is asymmetric, and the triangle inequality is not guaranteed; we therefore call it a divergence rather than a distance.
In dictionary learning and assignment, $C_j$ is the dictionary element.
We call the method that applies this divergence to normalized curves \zPD{}.

\zPD{} incorporates amplitude normalization, time warping, rotation, scaling, translation, and vertical flipping, and regularization keeps rotation, scaling, and translation close to the identity.
The same method without normalization is called ProDTW. \rev{ProDTW lacks independent scaling of the value axis because $\rho$ scales both axes equally.}

\subsection{Dictionary Construction}\label{sec:dictionary}
Train candidate patterns are divided into $K$ groups of similar shapes; the set of candidate patterns representing the groups, $\mathcal Q=\{Q_1,\dots,Q_K\}$, is called the dictionary, and each $Q_k$ a dictionary element.
Dictionary elements are not synthesized curves but actual observed candidate patterns $C_i$.
They are chosen to reduce the sum of squared divergences from each candidate pattern to its nearest element,
\begin{equation}
\mathcal L(\mathcal Q)=\sum_i\min_{1\le k\le K}D(C_i,Q_k)^2
\label{eq:kmedoids}
\end{equation}
(k-medoids~\cite{kaufman1990finding}).
After choosing initial elements from mutually dissimilar candidate patterns, we repeat two steps: assigning each candidate pattern to the group of its nearest element, and replacing each element with the candidate pattern in its group that minimizes the sum of squares.
To reduce computation, the second step tries only a random subset of candidate patterns and evaluates the sum on a subset, but the current element is always included in the comparison and is replaced only when the sum decreases.
Because elements are observed curves, they can be traced back to their positions in the original series.

\begin{table}[tb]
\caption{Main settings}
\label{tab:settings}
\centering
\small
\setlength{\tabcolsep}{3pt}
\begin{tabular}{lp{0.62\columnwidth}}
\toprule
Item & Setting \\
\midrule
Timeframes & 1m, 5m, 15m, 30m, 1h, 4h, 1D \\
Window length (obs.) & 1m: 30--960; 5m: 12--384; 15m: 12--192; 30m, 1h: 12--96; 4h, 1D: 8--64 \\
Curves & $M=64$; quadratic Savitzky--Golay, window ratio 0.08 \\
DTW & Sakoe--Chiba radius $w=6$; $\le$10 alternating iterations \\
Regularization & $(\lambda_s,\lambda_r,\lambda_t,\lambda_\pi)=(0.05,0.30,0.05,0.02)$ \\
Dictionary & $K\in\{5,10,15,20,30,50\}$; seeds 1, 2, 3; element updates evaluated on subsets of candidate patterns \\
Occurrence & $\delta$: 95th percentile of the minimum divergence on Validation; $m_{\min}=0.05$ \\
\bottomrule
\end{tabular}
\end{table}

\begin{figure*}[!t]
\centering
\begin{minipage}[t]{0.49\textwidth}\centering
\includegraphics[width=\linewidth]{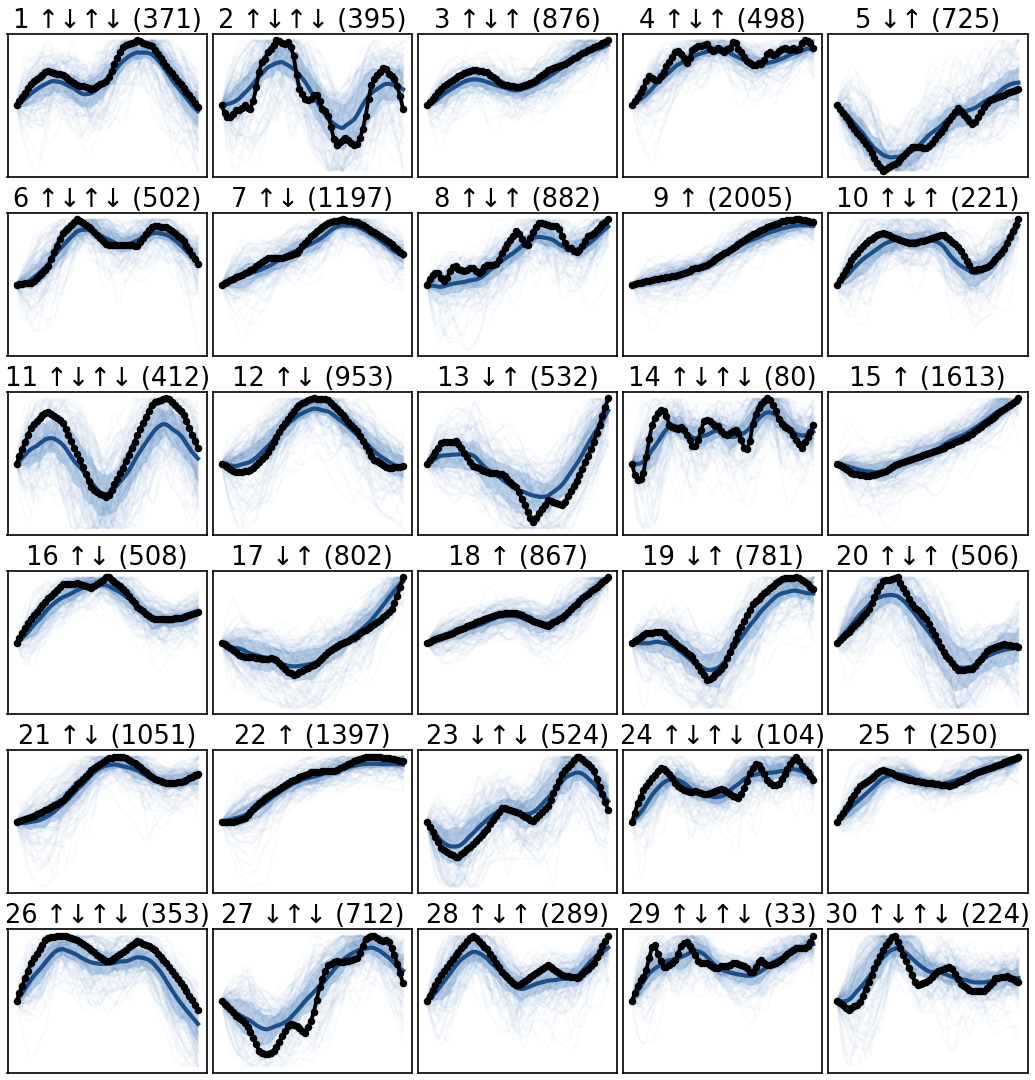}\\(a) FTSE
\end{minipage}\hfill
\begin{minipage}[t]{0.49\textwidth}\centering
\includegraphics[width=\linewidth]{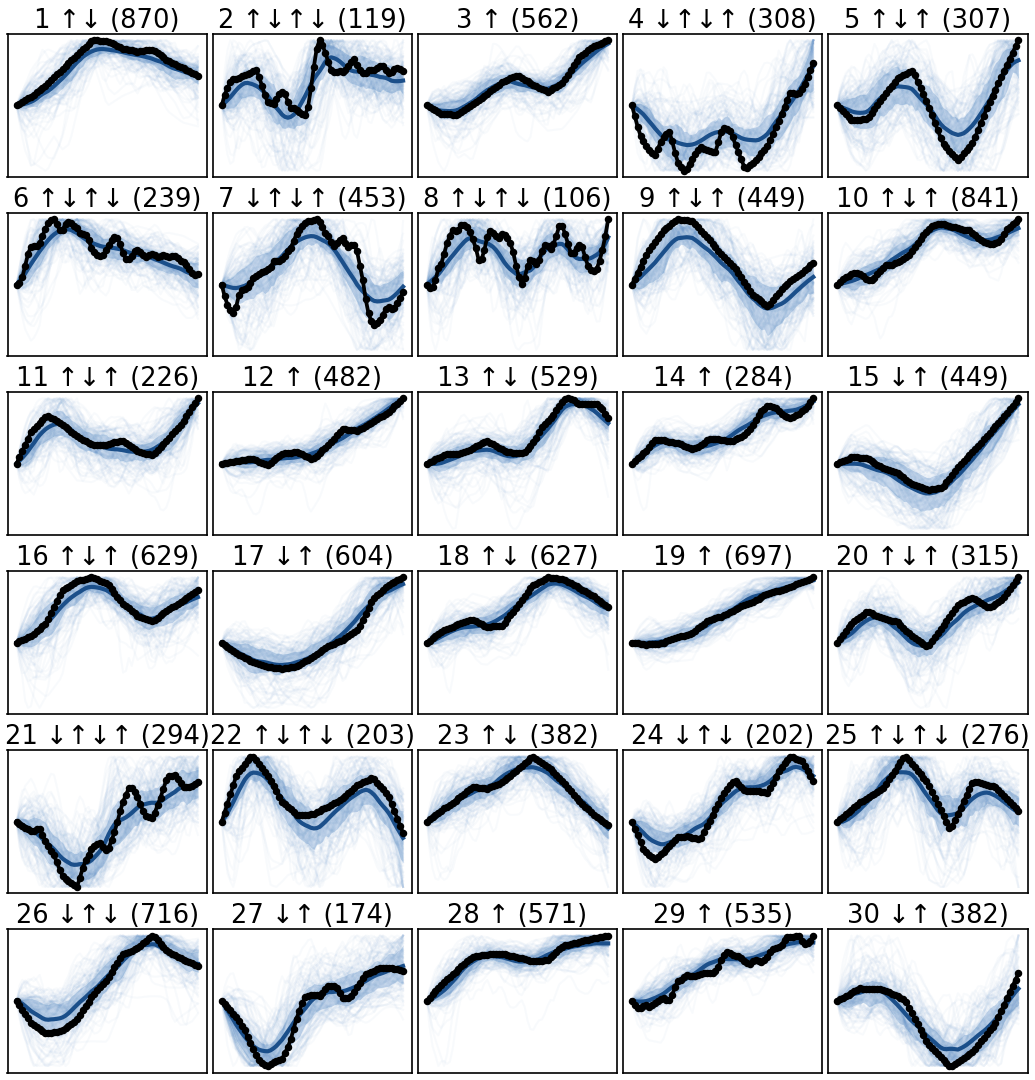}\\(b) Nikkei
\end{minipage}
\caption{Dictionaries of 30 shapes learned by \zPD{} ($K=30$, seed 1). Black lines are dictionary elements (observed curves); light blue lines are a subset of windows judged as occurrences of the element (orientation aligned and normalized by maximum absolute value); dark blue lines are their mean, and bands the interquartile range. Each title gives the element number, the sequence of rises ($\uparrow$) and falls ($\downarrow$), and in parentheses the number of occurrences in Test.}
\label{fig:dictionary}
\end{figure*}

\section{Experimental Design}\label{sec:design}

We evaluate the dictionary from three viewpoints: how accurately it reconstructs an unseen period with few elements (reconstruction accuracy; the main evaluation), whether its elements are mutually different shapes (diversity), and whether the obtained shapes are specific to market data (market specificity).

\subsection{Compared Methods}\label{sec:methods}
\rev{The six methods share candidate-pattern extraction, k-medoids dictionary construction, and the occurrence rule (Table~\ref{tab:methods}). With vertical flipping fixed, the four DTW-based methods form a $2\times2$ comparison of amplitude normalization and regularized Procrustes superimposition. Among normalized methods, normED+inv, normSBD+inv, normDTW+inv, and \zPD{} compare fixed correspondence, a uniform temporal shift, local time warping, and additional regularized superimposition, respectively; the two designs share normDTW+inv and \zPD{}, yielding six distinct methods. The DTW-based baselines use the same planar points, band, and time-offset penalty $\lambda_\pi$ as \zPD{} with the transform fixed to the identity, but their distance includes the penalty. For normalized values $x=(\tilde c_{i,m})_m$ and $y=(\tilde c_{j,m})_m$, normED+inv uses $\min_{o}\bigl(\frac{1}{M}\sum_m(x_m-o\,y_m)^2\bigr)^{1/2}$, and normSBD+inv uses a square-root form of the shape-based distance (SBD) of k-Shape~\cite{paparrizos2015kshape}, $\min_{o}\bigl(1-\max_h\mathrm{NCC}_h(x,o\,y)\bigr)^{1/2}$, where $\mathrm{NCC}_h$ is the normalized cross-correlation at shift $h$, taken over all shifts with zero padding. Representatives remain observed curves rather than the synthesized centroids of k-means or k-Shape, so these comparisons assess the handling of deformations within a common pipeline, not clustering algorithms or independent effects of individual transformation parameters.}

\begin{table}[tb]
\caption{Degrees of freedom of the compared methods and the differences they absorb. Norm.: amplitude (level and range); align.: timing (local time warping by DTW, uniform shift, or fixed correspondence); rot./scale: slope, size, and position (regularized rotation, scaling, and translation); flip: vertical orientation}
\label{tab:methods}
\centering
\small
\setlength{\tabcolsep}{4pt}
\begin{tabular}{lcccc}
\toprule
Method & Norm. & Align. & Rot./scale & Flip \\
\midrule
\B{\zPD{}} (proposed) & \checkmark & DTW & \checkmark & \checkmark \\
ProDTW & & DTW & \checkmark & \checkmark \\
normDTW+inv & \checkmark & DTW & & \checkmark \\
DTW+inv & & DTW & & \checkmark \\
\rev{normSBD+inv} & \checkmark & shift & & \checkmark \\
\rev{normED+inv} & \checkmark & fixed & & \checkmark \\
\bottomrule
\end{tabular}
\end{table}

\begin{figure*}[!t]
\centering
\includegraphics[width=\textwidth]{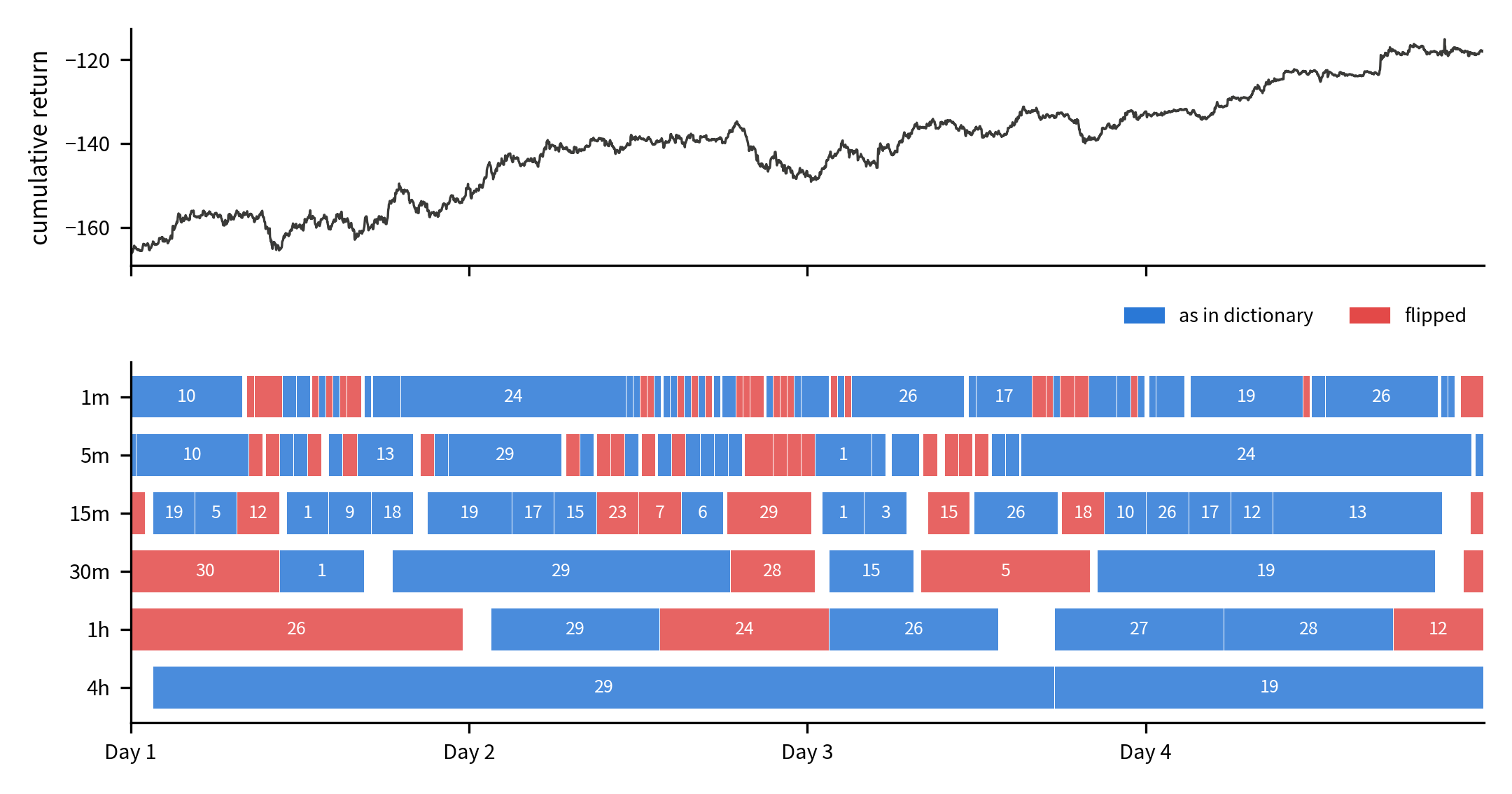}
\caption{Per-timeframe assignment over four Test days (Nikkei, \zPD{}, $K=30$, seed 1; \rev{a day denotes 1,440 one-minute observations}). Top: cumulative market return. Bottom: each row shows the occurrences of dictionary elements at that timeframe in time order. Colors indicate orientation (blue: as in the dictionary; red: flipped), and labels are element numbers (shared with Fig.~\ref{fig:dictionary}(b)).}
\label{fig:lanes}
\end{figure*}

\subsection{Assigning the Dictionary}\label{sec:recon}
The dictionary is learned exclusively from the training set and assigned to all windows extracted with sliding start times from Train, Validation, and Test.
For each window, we compute the divergence $d_1$ to the nearest element $k^\ast$ and $d_2$ to the second nearest; if $d_1\le\delta$ and the margin $(d_2-d_1)/d_2\ge m_{\min}$, element $k^\ast$ is judged to \emph{occur} in the window.
That is, a window is regarded as the same shape only when it is close enough to element $k^\ast$ and clearly closer than to the second element.
The threshold $\delta$ is fixed at the 95th percentile of the minimum divergence on Validation, and Test is used neither for learning nor for setting the threshold.

A window judged as an occurrence can be reconstructed in the original units of the market series by applying the correspondence $\pi$ and transform $T_{o,\theta}$ to the element and restoring $\bar c_i$ and $s_i$ removed by normalization.
Because windows overlap in time, for each timeframe we give larger weights to occurrences with longer windows and smaller divergences, and select the set of non-overlapping occurrences maximizing the total weight exactly by weighted interval scheduling~\cite{kleinberg2006algorithm}.
Arranged in time order, this set is called the assignment, and it represents the price series as a sequence of dictionary elements at that timeframe (Fig.~\ref{fig:lanes}).

\subsection{Evaluation Metrics}\label{sec:metrics}
\subsubsection{Reconstruction accuracy}\label{sec:m-compact}
For each method we learn dictionaries for $K\in\{5,10,15,20,30,50\}$ and three seeds, assign every Test window to its nearest element, and compute two reconstruction scores.
The overlapping $R^2$ is computed by reconstructing all overlapping windows, using each window's mean as the reference.
The non-overlapping $R'^2$ is obtained by joining reconstructions of non-overlapping windows at each timeframe to reconstruct the entire Test period, as
\begin{equation}
R'^2=1-\frac{\sum_t(x_t-\hat x_t)^2}{\sum_t(x_t-\ell_t)^2},
\label{eq:skill}
\end{equation}
where $x_t$ is the cumulative market return at 1-minute resolution, $\hat x_t$ its reconstruction, and $\ell_t$ a piecewise linear approximation with the same number of segments~\cite{keogh2001online}.
Windows whose error is smaller than that of a straight line are selected by the same method as in Section~\ref{sec:recon}, and uncovered intervals are filled with straight lines.
To examine whether differences persist when the reconstruction transform is restricted, both metrics are also computed under a condition that allows all methods only the same linear map $a\,Q_k+b$ of element $Q_k$ (common transform). \rev{In both conditions, each method uses its own point correspondence: the DTW path, the shift that maximizes the cross-correlation (with zero padding) for normSBD+inv, and the identity correspondence for normED+inv.}

\subsubsection{Diversity}\label{sec:m-redundancy}
The inter-element distance is the median, over elements, of the distance to the nearest other element, measured by the normDTW+inv distance common to all methods.
The effective rank~\cite{roy2007effective} indicates how many independent directions the matrix of normalized elements effectively spans.
The former looks at pairwise closeness under time warping, the latter at the linear spread of the whole dictionary.

\subsubsection{Market specificity}\label{sec:m-null}
We replace the market series with two kinds of surrogate series~\cite{schreiber2000surrogate}, repeat the same procedure \rev{for \zPD{} and DTW+inv}, and compare how evenly elements are used in Test and the combinations of adjacent elements and of occurrences across timeframes.
The surrogates are a random permutation of the returns and \rev{an amplitude-adjusted Fourier-transform (AAFT) surrogate}~\cite{theiler1992testing} that preserves the return distribution and approximately the autocorrelation of the market series. \rev{Transitions between adjacent occurrences are tested against independence of consecutive elements with a Poisson upper-tail test and the Benjamini--Hochberg correction~\cite{benjamini1995controlling} ($q\le0.05$).}

For reconstruction accuracy and diversity, methods are paired for each combination of market, $K$, and seed, and we report the mean of the 36 paired differences and the Wilcoxon signed-rank test~\cite{wilcoxon1945individual}.
Since the 36 pairs evaluate the same Test period repeatedly and are not independent samples, $p$-values are treated as reference values, and consistency is judged by whether the sign of the difference agrees across conditions.

\section{Results}\label{sec:results}

\begin{table}[tb]
\caption{Per-timeframe assignment in Test ($K=30$, seed 1): coverage (fraction of the Test period covered by occurrences), occurrences per 1,440 one-minute observations, and number of distinct elements used}
\label{tab:lanes}
\centering
\small
\setlength{\tabcolsep}{3pt}
\begin{tabular}{llcccccc}
\toprule
 & & 1m & 5m & 15m & 30m & 1h & 4h \\
\midrule
FTSE & Coverage & .97 & .95 & .95 & .94 & .91 & .92 \\
 & Occurrences & 17.0 & 12.3 & 4.6 & 2.4 & 1.2 & 0.4 \\
 & Distinct & 27 & 25 & 21 & 18 & 13 & 9 \\
\midrule
Nikkei & Coverage & .97 & .96 & .95 & .93 & .96 & .88 \\
 & Occurrences & 13.5 & 10.7 & 5.0 & 2.5 & 1.3 & 0.6 \\
 & Distinct & 27 & 25 & 24 & 18 & 14 & 6 \\
\bottomrule
\end{tabular}
\end{table}

\begin{table*}[!t]
\caption{Reconstruction accuracy and diversity by dictionary size $K$ (mean over three seeds) and paired comparisons with \zPD{} (36 pairs: 2 markets $\times$ 6 values of $K$ $\times$ 3 seeds). ``Diff.'' is the mean of \zPD{} minus each method; $p$ is from the Wilcoxon signed-rank test (reference only, as conditions are not independent). Larger is better for all metrics. Bold marks the best value in each $K$ column and, for Diff.\ and $p$, comparisons in which \zPD{} is higher with $p<0.05$ (reference).}
\label{tab:eval}
\centering
\footnotesize
\setlength{\tabcolsep}{2.1pt}
\begin{tabular}{lcccccccccccccc}
\toprule
 & \multicolumn{6}{c}{FTSE} & \multicolumn{6}{c}{Nikkei} & \multicolumn{2}{c}{Paired comparison} \\
\cmidrule(lr){2-7}\cmidrule(lr){8-13}\cmidrule(lr){14-15}
$K$ & 5 & 10 & 15 & 20 & 30 & 50 & 5 & 10 & 15 & 20 & 30 & 50 & Diff. & $p$ \\
\midrule
\multicolumn{15}{l}{\textit{Overlapping $R^2$}} \\
\quad \zPD{} & \B{.909} & \B{.937} & \B{.947} & \B{.953} & \B{.961} & \B{.966} & \B{.902} & \B{.925} & \B{.942} & \B{.949} & \B{.956} & \B{.962} &  &  \\
\quad ProDTW & .902 & .922 & .933 & .940 & .948 & .956 & .889 & .914 & .922 & .930 & .935 & .943 & \B{+.015} & \B{$<$1e-6} \\
\quad normDTW+inv & .901 & .923 & .938 & .945 & .953 & .960 & .888 & .919 & .934 & .940 & .950 & .958 & \B{+.008} & \B{$<$1e-6} \\
\quad DTW+inv & .596 & .668 & .703 & .749 & .798 & .833 & .441 & .621 & .653 & .698 & .736 & .805 & \B{+.251} & \B{$<$1e-6} \\
\quad \rev{normSBD+inv} & .805 & .823 & .839 & .851 & .861 & .875 & .792 & .820 & .829 & .839 & .849 & .862 & \B{+.105} & \B{$<$1e-6} \\
\quad \rev{normED+inv} & .746 & .793 & .818 & .828 & .840 & .856 & .733 & .777 & .795 & .807 & .827 & .842 & \B{+.137} & \B{$<$1e-6} \\
\midrule
\multicolumn{15}{l}{\textit{Non-overlapping $R'^2$}} \\
\quad \zPD{} & .741 & \B{.823} & \B{.841} & \B{.848} & \B{.853} & \B{.862} & .706 & \B{.767} & \B{.796} & \B{.812} & \B{.821} & \B{.826} &  &  \\
\quad ProDTW & .751 & .779 & .786 & .802 & .815 & .823 & .688 & .762 & .776 & .789 & .797 & .816 & \B{+.026} & \B{$<$1e-6} \\
\quad normDTW+inv & \B{.760} & .785 & .812 & .821 & .834 & .844 & \B{.724} & .748 & .775 & .775 & .801 & .818 & \B{+.016} & \B{4e-4} \\
\quad DTW+inv & .082 & .281 & .320 & .433 & .368 & .464 & .352 & .459 & .543 & .527 & .603 & .665 & \B{+.383} & \B{$<$1e-6} \\
\quad \rev{normSBD+inv} & .536 & .590 & .622 & .630 & .641 & .678 & .533 & .559 & .576 & .579 & .621 & .611 & \B{+.210} & \B{$<$1e-6} \\
\quad \rev{normED+inv} & .491 & .565 & .582 & .596 & .599 & .645 & .440 & .525 & .564 & .584 & .594 & .612 & \B{+.241} & \B{$<$1e-6} \\
\midrule
\multicolumn{15}{l}{\textit{Inter-element distance}} \\
\quad \zPD{} & \B{.632} & \B{.450} & .394 & \B{.362} & .298 & .269 & .574 & \B{.488} & \B{.396} & \B{.349} & .302 & .268 &  &  \\
\quad ProDTW & .409 & .337 & .304 & .279 & .254 & .251 & .485 & .371 & .357 & .279 & .273 & .236 & \B{+.079} & \B{$<$1e-6} \\
\quad normDTW+inv & .578 & .432 & \B{.397} & .350 & \B{.309} & \B{.280} & .498 & .376 & .343 & .306 & .270 & .260 & \B{+.032} & \B{5e-4} \\
\quad DTW+inv & .229 & .261 & .256 & .207 & .204 & .184 & .280 & .261 & .224 & .231 & .237 & .210 & \B{+.166} & \B{$<$1e-6} \\
\quad \rev{normSBD+inv} & .501 & .391 & .343 & .329 & .276 & .243 & .487 & .337 & .358 & .346 & \B{.303} & \B{.270} & \B{+.050} & \B{1e-4} \\
\quad \rev{normED+inv} & .547 & .356 & .303 & .287 & .262 & .248 & \B{.585} & .412 & .390 & .343 & .302 & .265 & \B{+.040} & \B{3e-4} \\
\midrule
\multicolumn{15}{l}{\textit{Effective rank}} \\
\quad \zPD{} & \B{3.27} & 3.83 & 4.62 & 4.68 & 5.07 & 5.53 & \B{3.41} & \B{4.44} & \B{4.52} & \B{4.61} & 4.91 & \B{5.92} &  &  \\
\quad ProDTW & 2.49 & 3.52 & 3.70 & 3.97 & 4.32 & 5.10 & 2.55 & 3.57 & 4.16 & 4.21 & 4.81 & 5.28 & \B{+0.59} & \B{$<$1e-6} \\
\quad normDTW+inv & 2.96 & \B{4.35} & \B{4.66} & \B{4.81} & \B{5.27} & \B{5.61} & 3.05 & 4.02 & 4.21 & 4.60 & 4.75 & 5.39 & +0.09 & .16 \\
\quad DTW+inv & 2.13 & 2.80 & 3.48 & 3.54 & 3.54 & 4.12 & 2.09 & 2.97 & 3.80 & 4.08 & 4.36 & 4.63 & \B{+1.11} & \B{$<$1e-6} \\
\quad \rev{normSBD+inv} & 2.66 & 3.80 & 4.24 & 4.34 & 4.80 & 5.11 & 2.85 & 3.52 & 4.20 & 4.50 & \B{5.00} & 5.53 & \B{+0.35} & \B{2e-4} \\
\quad \rev{normED+inv} & 3.14 & 3.69 & 3.81 & 4.01 & 4.75 & 5.08 & 2.97 & 3.86 & 4.36 & 4.57 & 4.74 & 5.47 & \B{+0.36} & \B{1e-4} \\
\bottomrule
\end{tabular}
\end{table*}

\subsection{Learned Dictionaries}\label{sec:res-dict}
Fig.~\ref{fig:dictionary} shows the 30 patterns learned by \zPD{} for the UK and Japanese markets.
Besides monotone moves, the dictionaries contain reversals, rebounds after pullbacks, and multiple turns, and all 30 elements were observed in Test in both markets.

Fig.~\ref{fig:lanes} shows the assignment of this dictionary at each timeframe over four volatile days of the Nikkei Test period.
Pattern occurrences are denser and shorter at finer timeframes and sparser and longer at coarser ones, so the same price movement is represented as sequences of shapes at different granularities.
At the 1-minute timeframe, \rev{1,440 observations are represented by about} 14--17 occurrences (Table~\ref{tab:lanes}).
\texttt{1D} is excluded from Fig.~\ref{fig:lanes} and Table~\ref{tab:lanes} because the Test period spans only about 28 (FTSE) and 18 (Nikkei) units of 1,440 observations, yielding very few windows.

\begin{figure*}[!t]
\centering
\includegraphics[width=\textwidth]{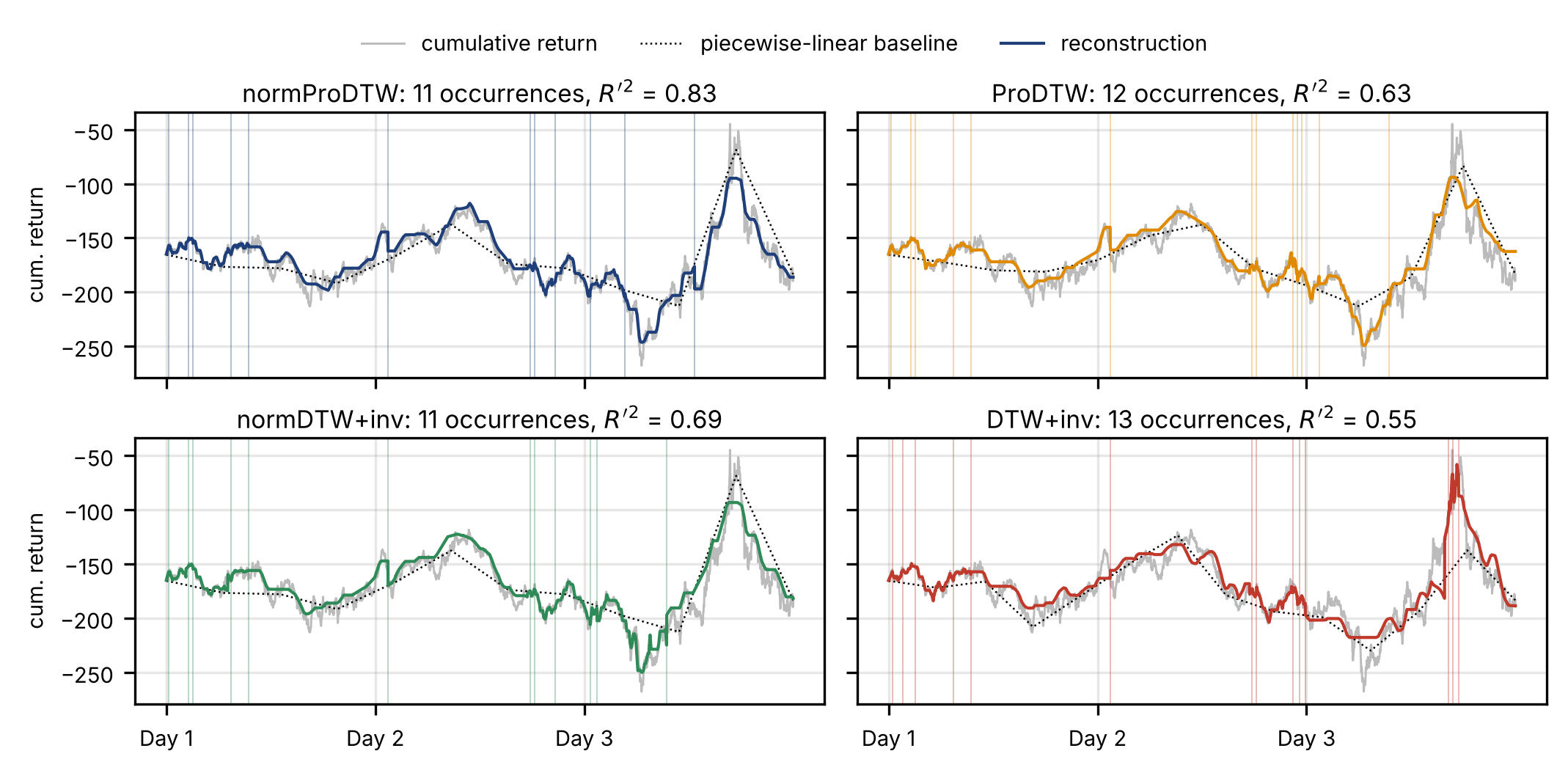}
\caption{Reconstruction of the first three Test days (FTSE, $K=10$, seed 1, 1-minute timeframe; \rev{days as in Fig.~\ref{fig:lanes}}). Gray: cumulative return; dotted: piecewise linear approximation with the same number of segments; colored: \rev{reconstruction by the four DTW-based methods}; vertical lines: occurrence start times. Titles give the number of occurrences and $R'^2$ in this period.}
\label{fig:recon}
\end{figure*}

\subsection{Evaluation of the Dictionaries}\label{sec:res-eval}
Table~\ref{tab:eval} summarizes the results.

\begin{table*}[!t]
\caption{Test coverage and $R^2$ for the four DTW-based methods under a common criterion (normalized reconstruction error $\le\varepsilon$), averaged over timeframes and three seeds. Bold marks the best value in each row for each metric.}
\label{tab:tradeoff}
\centering
\footnotesize
\setlength{\tabcolsep}{4pt}
\begin{tabular}{lcccccccccc}
\toprule
 & & & \multicolumn{4}{c}{Coverage} & \multicolumn{4}{c}{$R^2$} \\
\cmidrule(lr){4-7}\cmidrule(lr){8-11}
Market & $K$ & $\varepsilon$ & \zPD{} & ProDTW & normDTW+inv & DTW+inv & \zPD{} & ProDTW & normDTW+inv & DTW+inv \\
\midrule
FTSE & 10 & 0.05 & \B{.713} & .590 & .642 & .160 & \B{.991} & .988 & .989 & .984 \\
 & & 0.10 & \B{.892} & .870 & .857 & .593 & \B{.985} & .978 & .983 & .955 \\
 & & 0.20 & \B{.953} & .950 & .934 & .820 & \B{.980} & .971 & .974 & .919 \\
 & 30 & 0.05 & \B{.791} & .713 & .760 & .286 & \B{.994} & .990 & .992 & .985 \\
 & & 0.10 & \B{.936} & .897 & .898 & .725 & \B{.990} & .984 & .988 & .964 \\
 & & 0.20 & \B{.961} & .958 & .957 & .918 & \B{.988} & .979 & .983 & .936 \\
\midrule
Nikkei & 10 & 0.05 & \B{.685} & .612 & .606 & .117 & \B{.991} & .990 & .990 & .986 \\
 & & 0.10 & .860 & \B{.892} & \B{.892} & .593 & \B{.985} & .981 & .983 & .950 \\
 & & 0.20 & \B{.951} & .946 & \B{.951} & .859 & .976 & \B{.977} & \B{.977} & .911 \\
 & 30 & 0.05 & \B{.803} & .681 & .802 & .208 & \B{.993} & .991 & .992 & .986 \\
 & & 0.10 & \B{.929} & .913 & .922 & .694 & \B{.990} & .986 & .988 & .963 \\
 & & 0.20 & \B{.956} & .955 & .954 & .913 & \B{.988} & .983 & .986 & .940 \\
\bottomrule
\end{tabular}
\end{table*}

\subsubsection{Reconstruction accuracy}\label{sec:res-compact}
\zPD{} exceeded all \rev{other} methods on average in both $R^2$ and $R'^2$; the number of conditions out of 36 in which it was higher was 35--36 for $R^2$ and 29--36 for $R'^2$.
The gap is particularly large against DTW+inv, which tends to treat shapes with different ranges as different elements; \zPD{} at $K=5$ exceeds DTW+inv at $K=50$ \rev{in both markets and both metrics}.
The gap to the strongest baseline, normDTW+inv, is small but consistent (\rev{for seed means, }$R'^2$ is higher in all 10 \rev{market--size combinations} with $K\ge10$ and lower in the 2 with $K=5$), and \zPD{} reaches the $R'^2$ that normDTW+inv attains at $K=50$ with \rev{approximately} 17--28 elements\rev{, obtained by linear interpolation between the evaluated sizes}.
The gap remains under the common transform (differences from normDTW+inv of $+0.005$ in $R^2$ and $+0.009$ in $R'^2$, higher in 34 and 26 of 36 conditions), so the improvement cannot be explained by the freedom of the transform allowed at reconstruction alone, and the difference in the selected elements appears to contribute.
Note that $R'^2$ also includes the straight-line filling of uncovered intervals and that the selected windows\rev{, and hence the piecewise-linear reference,} differ across methods, so the gain cannot be attributed entirely to the quality of the elements.
\rev{Against the measures outside the DTW family, \zPD{} was higher than normSBD+inv and normED+inv in all 36 conditions, by $+0.105$ and $+0.137$ in $R^2$ and by $+0.210$ and $+0.241$ in $R'^2$ on average; the gaps remained under the common transform ($+0.190$ and $+0.213$ in $R'^2$).}
\rev{Fig.~\ref{fig:recon} illustrates reconstruction of the first three FTSE Test days using the four DTW-based methods at $K=10$.}
\zPD{} follows the series even in volatile phases ($R'^2=0.83$ versus 0.55--0.69 for the three DTW-based comparison methods), whereas DTW+inv departs markedly from the series around the surge on the third day.

\subsubsection{Diversity}\label{sec:res-redundancy}
The inter-element distance of \zPD{} is larger on average than that of all other methods (higher in 33, 29, and 35 of 36 conditions than ProDTW, normDTW+inv, and DTW+inv, respectively, \rev{and in 28 of 36 than each of normSBD+inv and normED+inv}), so the elements remained distinct shapes despite the added degrees of freedom.
The effective rank differs little from normDTW+inv (higher in 24 of 36 conditions) and is at most about 6 at $K=50$ for every method; the main difference between the two therefore appears to lie in shape dissimilarity under time warping rather than in linear spread.

\subsubsection{Robustness to the occurrence criterion}\label{sec:res-tradeoff}
\rev{For the four DTW-based methods, we also examined robustness under a common criterion, varying the normalized reconstruction error} $\varepsilon$ relative to the window's range (Table~\ref{tab:tradeoff}).
\zPD{} was best in most conditions for both coverage (fraction of the Test period covered by windows meeting the criterion) and $R^2$ of those windows: for coverage it was uniquely best in 10 of 12 conditions and tied in 1, and for $R^2$ it was best in 11 of 12 (\rev{all exceptions occurred for Nikkei at $K=10$}). The ranking of these four methods was largely stable with respect to the choice of criterion.

\subsubsection{Market specificity}\label{sec:res-null}
Table~\ref{tab:null} shows how dictionary elements are used when learned on the market series and on the two surrogate series.
For \zPD{}, the effective number of elements and the fraction of elements occurring 10 or more times are \rev{broadly similar} across the three series; almost no transitions between adjacent occurrences are significant after multiple-comparison correction (only one pair at the Nikkei 15-minute timeframe), and links between occurrences across timeframes appear to a similar degree in the surrogates.
The usage of DTW+inv varies widely across series, but the market series is not consistently more skewed.
Hence, under the present metrics and surrogates, we found no evidence supporting shapes specific to the market; the dictionary should be understood as a general-purpose dictionary of shapes for describing price paths, and this study makes no claim of predictability.

\begin{table}[tb]
\caption{Usage of dictionary elements learned on the market series and on surrogate series ($K=30$, seed 1). The effective number of elements is the exponential of the entropy of the distribution of occurrence counts in Test; the fraction is the share of elements occurring 10 or more times in Test.}
\label{tab:null}
\centering
\footnotesize
\setlength{\tabcolsep}{2.5pt}
\begin{tabular}{llcccccc}
\toprule
 & & \multicolumn{3}{c}{Effective number} & \multicolumn{3}{c}{Fraction $\ge$10} \\
\cmidrule(lr){3-5}\cmidrule(lr){6-8}
Market & Method & Market & Perm. & Phase & Market & Perm. & Phase \\
\midrule
FTSE & \zPD{} & 23.8 & 25.3 & 24.5 & 1.00 & 1.00 & 1.00 \\
 & DTW+inv & 13.9 & 5.4 & 6.2 & 0.73 & 0.60 & 0.50 \\
\midrule
Nikkei & \zPD{} & 26.7 & 23.9 & 23.4 & 1.00 & 1.00 & 1.00 \\
 & DTW+inv & 2.7 & 6.3 & 6.6 & 0.37 & 0.47 & 0.57 \\
\bottomrule
\end{tabular}
\end{table}

\section{Discussion}\label{sec:discussion}

\rev{The comparisons in Table~\ref{tab:eval} suggest three findings about which differences should be absorbed. First, normalization improves mean $R'^2$ by approximately $0.367$ without regularized superimposition and $0.026$ with it; conversely, superimposition improves it by $0.357$ without normalization and $0.016$ with it. These diminishing gains suggest partly overlapping effects, particularly in accounting for differences in range, although the smaller remaining error may also contribute; normalization acts more directly because it adjusts the value axis independently of time.}

\rev{Second, mean $R'^2$ increases by about $0.032$ from normED+inv to normSBD+inv (higher in 26 of 36 conditions) and by $0.193$ from normSBD+inv to normDTW+inv, suggesting that local differences in the timing of rises, falls, and reversals matter more than a single temporal offset.}

\rev{Third, regularized superimposition improves mean $R'^2$ over normDTW+inv by $0.016$, or $0.009$ under the common transform, and increases mean inter-element distance by $0.032$, suggesting a benefit beyond normalization and time warping. These are conditional comparisons within the dictionary pipeline, in which alignment costs, selected elements, and windows also differ; they do not isolate rotation, scaling, translation, or vertical flipping.}

On the other hand, with a very small dictionary ($K=5$), \zPD{} \rev{has slightly lower non-overlapping $R'^2$ than normDTW+inv in both markets}.
When one element must represent different shapes, the penalties on the transform are likely to limit the fit.

\rev{Dictionary size was varied over six values (Table~\ref{tab:eval}), and robustness to alternative occurrence criteria was examined (Table~\ref{tab:tradeoff}). The comparison with normDTW+inv examines the contribution of regularized superimposition rather than sensitivity to the individual regularization weights. These weights, the DTW band, the set of window lengths, and the settings for occurrence thresholds (the percentile for $\delta$ and the margin $m_{\min}$) were not systematically varied.}

\rev{Reconstruction accuracy and diversity do not establish forecasting performance or economic value, which must be evaluated on specific tasks. Both datasets cover the 2007--2009 global financial crisis, and the comparisons share one Test period per market rather than independent replications. Generalization to calmer or recent periods, individual stocks, other asset classes, and transfer of a dictionary across markets remain to be evaluated.}

\section{Conclusion}\label{sec:conclusion}
\rev{We proposed \zPD{} to construct compact multiscale dictionaries of observed financial time-series shapes. On FTSE- and Nikkei-based market series, the dictionaries achieved higher average reconstruction accuracy and inter-element dissimilarity than the compared methods. Amplitude normalization and local time warping contributed substantially, while regularized Procrustes superimposition provided a smaller additional gain.}

\rev{Each dictionary element could be annotated with a textual description of its sequence of rises and falls. The assignment and the attributes of each occurrence could then be passed to a language model to generate commentary on price movements across timeframes, possibly combined with news. Evaluating such commentary and applications such as retrieval-augmented forecasting~\cite{han2025raf} remains future work.}



\end{document}